\pdfoutput=1
\documentclass[11pt]{article}
\usepackage{makecell}

\usepackage{acl}

\usepackage{times}
\usepackage{latexsym}

\usepackage[T1]{fontenc}
\usepackage[utf8]{inputenc}

\usepackage{microtype}

\usepackage{inconsolata}

\usepackage{caption}
\usepackage[separate-uncertainty=true]{siunitx}
\usepackage{mathtools}
\usepackage{graphicx}
\usepackage{courier}
\usepackage{epstopdf}
\usepackage{color}
\usepackage{csquotes}

\usepackage{amsmath}
\usepackage{amsfonts}
\usepackage{amssymb}
\usepackage[subnum]{cases}
\usepackage{booktabs} %For formal tables
\usepackage{pifont}

\usepackage{algorithm,algorithmicx,algpseudocode}

\usepackage{amssymb}% http://ctan.org/pkg/amssymb
\usepackage{pifont}% http://ctan.org/pkg/pifont

\usepackage {tikz}
\usetikzlibrary {positioning}
\definecolor {processblue}{cmyk}{0.96,0,0,0}
\usepackage{lipsum,adjustbox}

\usepackage[subnum]{cases}
\usepackage{color,soul}

\usepackage{subcaption}

\usepackage{footnote}
\makesavenoteenv{tabular} %ICDE version does not support

\usepackage{multirow}

\usepackage{booktabs,subcaption,amsfonts,dcolumn}
\newcolumntype{d}[1]{D..{#1}}
\usepackage{xcolor}

\usepackage{textcomp}

\usepackage{bbold}
\usepackage{mathtools}
\usepackage{breqn}

\usepackage{amsthm}

\usepackage[many]{tcolorbox}
\newtcolorbox{boxA}{
    fontupper = \bf,
    boxrule = 1.5pt,
    colframe = black % frame color
}

\title{Enhancing Presentation Slide Generation by LLMs with \\a Multi-Staged End-to-End Approach}

\author{Sambaran Bandyopadhyay, Himanshu Maheshwari, Anandhavelu Natarajan, Apoorv Saxena \\
   Adobe Research \hspace{0.2cm} \\
  \texttt{\{sambaranb, himahesh, anandvn, apoorvs\}@adobe.com}\\
}

\begin{document}
\maketitle
\begin{abstract}
Generating presentation slides from a long document with multimodal elements such as text and images is an important task. This is time consuming and needs domain expertise if done manually. Existing approaches for generating a rich presentation from a document are often semi-automatic or only put a flat summary into the slides ignoring the importance of a good narrative. In this paper, we address this research gap by proposing a multi-staged end-to-end model which uses a combination of LLM and VLM. We have experimentally shown that compared to applying LLMs directly with state-of-the-art prompting, our proposed multi-staged solution is better in terms of automated metrics and human evaluation.
\end{abstract}

\section{Introduction}\label{sec:intro}
% {\color{red}Sambaran}
Presentations are a visually effective way to convey an idea to a broad audience \citep{bartsch2003effectiveness}. 
They are heavily used in academia, research, marketing and sales.
% and different other business to present a vast amount of content into a visually appealing and easy-to-follow format. 
A good presentation conveys a story (or a \textit{narrative}) in the form of a sequence of slides. It often needs to be generated from a long document which contains multimodal information such as text and images. However, making such a presentation from a document is very time expensive and needs domain expertise.
% \citep{piorkowski2021ai}.

There are rule-based approaches to generate a presentation from a document \citep{1565545,Winters2019AutomaticallyGE}.
Automatically generating a presentation from a given multimodal document is challenging because of several reasons. Compared to a flat document summary, the slide narrative should convey a story to its audience and is often non-linear with respect to the flow of information in the document \citep{hargood2009exploring}. The content of a slide should be concise, easy to follow and visually appealing. So, it needs reasoning over both text and images, and their inter-relationship. 
% There are some works which address some of these problems. 
Assuming the slide titles to be the same as the document sections, there are works which use a query specific summarizer \citet{sravanthi2009slidesgen}, learn sentence importance \citep{hu2013ppsgen} and extract hierarchical relations between phrases \citep{wang2017phrase} to generate the presentation.
\citet{sun2021d2s} takes the outline from the user and uses that to extract multimodal content and summarize that to slides. \citet{fu2022doc2ppt} proposes a sequence-to-sequence architecture and a trainable policy to determine when to proceed to the next section/slide. But, it needs large amount of document-to-slides parallel training data which makes it difficult to generalize and scale. 
% the approach not generalizable to multiple domains.

\begin{figure}
    \centering
    \includegraphics[width = \linewidth]{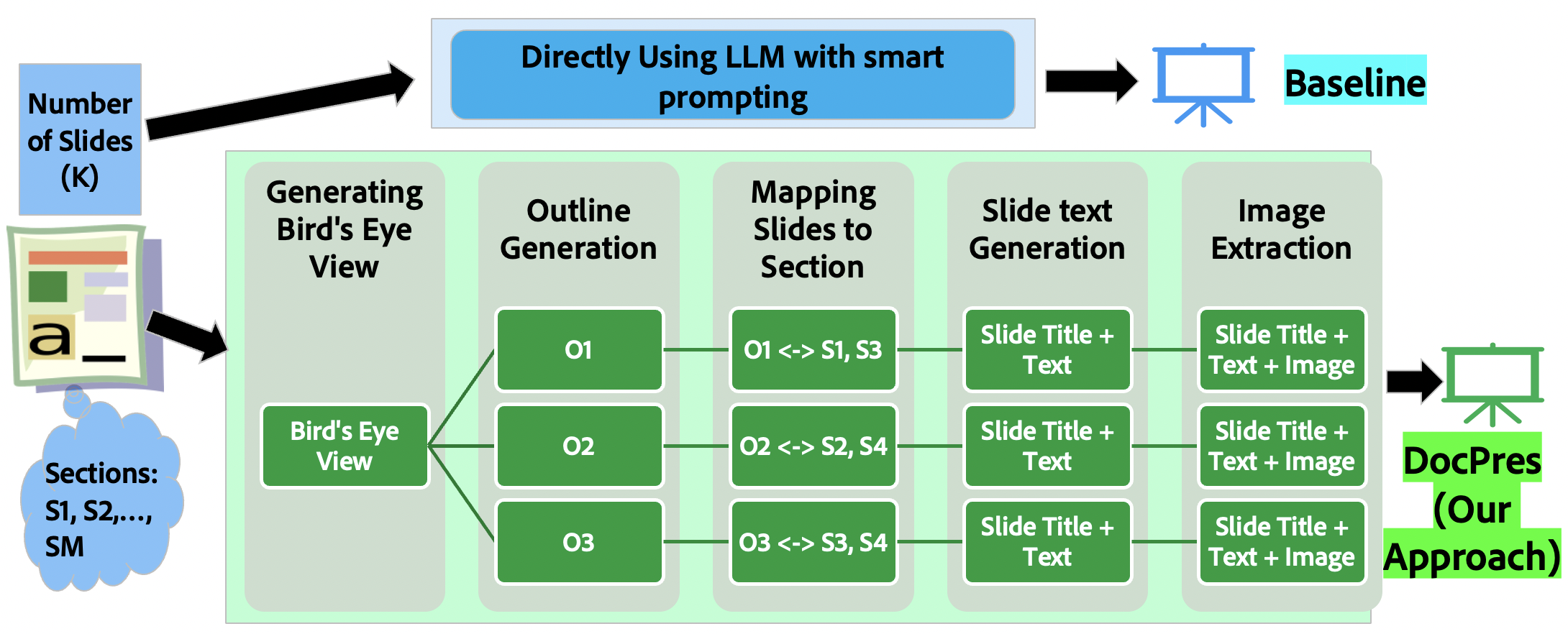} 
    \caption{Comparison of DocPres (in green) with a conventional way of generating a presentation directly using an LLM (in blue).}
    \label{fig:DocPres_archi}
\end{figure}

Recent developments in large language models (LLMs) and vision language models (VLMs) have been successfully applied in several multimodal generation tasks. These methods are also easy to use since they can generate content based on simple text prompts and can be generalized to multiple domains. However, compared to open domain generative task, generating a presentation from a specific document is much more challenging because of the following reasons: (i) It is difficult to feed an entire long document to an LLM because of its upper limit on the context length (the number of tokens it can process at a shot) \citep{mu2023learning}. 
% Existing approaches such as map reduce techniques \cite{topsakal2023creating}, fine-tuning an existing pre-trained transformer with a longer context window \citep{mu2023learning}, or using length extrapolation \citep{peng2023yarn} and position interpolation of transformers \citep{chen2023extending} are suboptimal, expensive and often infeasible to use for a black-box LLM like GPT-3.5. 
(ii) The performance of LLMs drops with the length of the context within a prompt. In fact the performance degrades significantly when LLMs need to access relevant information in the middle of long contexts  \cite{liu2023lost}, which is a must requirement for our task. 
(iii) Finally, LLMs are poor in attributing the exact source of the generated content (i.e., mapping a generated slide to some subsections of the document). Both VLMs and LLMs are prone to hallucinations and this tendency increases with the longer and incomplete context \citep{azamfirei2023large,zhou2023analyzing}. Thus, directly using LLMs for generating slides from a long document is not a good strategy. 

With this motivation, we try to divide the task of generating presentation from a long document into multiple simpler sub-tasks, each with a shorter context. Following are the novel contributions made in this paper:
\textbf{(i)} We have proposed an unsupervised multi-staged approach to generate slides from a long document, referred as DocPres (\textit{Doc}ument to \textit{Pres}entation). Our approach is multimodal-in and multimodal out in nature.
% i.e., both the input document and the output slides can have both text and images. 
\textbf{(ii)} Our proposed approach has multiple hierarchies and simpler stages, each with specific context as shown in Figure \ref{fig:DocPres_archi}.
% First, generating a bird's-eye view of the given document. Second, generating outlines (or the slide titles) from the bird's-eye view. Third, assign multiple sections of the given document to each point (slide title) in the outline. Next, paraphrases the corresponding text from the sections of the given document to generate the slides. Finally, extract images from the document to position them in the appropriate slides of the presentation. 
\textbf{(iii)} We conduct thorough experimentation involving a state-of-the-art LLM. We are able to show the merit of our multi-staged approach using different types of automated and human evaluated metrics.

% \section{Related Work}\label{sec:related}

% \section{Problem Formulation}\label{sec:probForm}
% Here, the input is a multimodal document which has text divided into multiple sections or subsections. The document can also have figures (images) present it it. For simplicity, we also refer charts, flow-diagrams, etc. as images in the document. The document also has a specific layout and we have the position (location co-ordinates in the page and the page numbers) for each element in it. More formally, the given document can be represented as $D = \{(S_i)_{i=1}^M, (F_j)_{j=1}^N)\}$, where $S_i$ is the $i$th section (we are using section and subsection exchangeably for simplicity for the rest of the paper) and $F_j$ is the $j$th figure in the document. A section $S_i = \{T_i, P_i\}$ is consisting of text and position (bounding box coordinates and page numbers) in the document. Similarly, a figure $F_j = \{{I_j, P_j}\}$ is consisting of the image and the position of it. Given the document, we aim to generate a set of slides $L = \{L_1,\cdots,L_K\}$ where each slide has some text and optional images coming from the information present in the document. The number of slides $K$ can be an optional input from the user and can be generated from our approach otherwise. The slides should be able to capture the key concepts present in the document, ensure a overall flow of the presentation, minimize any repetition of information and align multimodal content in it for the ease of understanding.

\section{Proposed Solution Approach}\label{sec:soln}
Let the input document be represented as $D = \{(S_i)_{i=1}^M, (F_j)_{j=1}^N)\}$, where $S_i$ is the $i$th section (or a subsection) and $F_j$ is the $j$th figure in the document. 
% A section $S_i = \{T_i, P_i\}$ is consisting of text and position (bounding box coordinates and page numbers) in the document. Similarly, a figure $F_j = \{{I_j, P_j}\}$ is consisting of the image and the position of it. 
Both sections and images are associated with their positions (bounding box coordinates and page numbers) in the document.
Given the document, we aim to generate a set of slides $L = \{L_1,\cdots,L_K\}$ where each slide has some text and optional images coming from the document.
% The number of slides $K$ can be an optional input from the user and can be generated otherwise. 

As the first step, we extract the text and images from the input document (pdf) using a publicly available API~\footnote{\url{https://developer.adobe.com/document-services/apis/pdf-extract/}} which gives the content of the document in a hierarchical fashion with section titles and the corresponding text within them with positions. Images present in the document are also extracted with their locations. 

\subsection{A Bird's-eye View of the Document}\label{sec:bird's-eye}
A bird's-eye view of a document refers to its hierarchical summary with sections, sub-sections and content within them. The sub-section is initially summarized, and these summaries are used to summarize the section. Consequently, the section summaries are used to summarise the entire document. We use an LLM to generate these summaries.
% Any LLM can create these summaries; we use GPT 3.5. 
This hierarchical approach ensures a layered and comprehensive overview of the document's content.

%The hierarchical summary is to a maximum depth of sub-sections. Beyond sub-sections depth, the content is treated as regular text, without further hierarchical organization. Initially, each sub-section is independently summarised by an LLM. Subsequently, sections are summarized by incorporating the summaries of their respective sub-sections and additional relevant text that was not part of any sub-sub-section. Lastly, the entire document is summarised using all the section summaries. This module's output is the document's hierarchical summary to the depth of the sub-section.

\subsection{Outline Generation}\label{sec:outline}
% {\color{red}Sambaran}
Here, we define the outline of the presentation as the sequence of the slide titles. Outline is important to control the high-level flow of information and convey the story from the document to a broader audience. Feeding the entire document to an LLM has two major drawbacks: limit on the context length of LLMs and their performance drop with the longer context as discussed in Section \ref{sec:intro}. So, we use the generated bird's-eye view of the document as the context and ask an LLM to generate $K$ important topics with a nice flow and short titles through a chain-of-thought prompt \cite{wei2022chain} (shown in Appendix \ref{sec:appendix_outline}).
% \begin{boxA}
% Think step by step.
% \begin{enumerate}
%     \item From the following text which contains a set of headings and some text within each heading, extract the most important <K> topics present in it
%     \item Length of each extracted topic should be small
%     \item The sequence of extracted topics should make a nice flow in a presentation
% \end{enumerate}
% <Bird's-eye view>
% \end{boxA}
% 
The output of the above call is the outline of the presentation as $O = \{O_1,\cdots,O_K\}$, where $O_k$ is the $k$-th slide title.

\subsection{Mapping Slides to Sections}\label{sec:slide2sec}
% {\color{red}Sambaran}
Once we obtain the outline of the presentation, the next task it to generate content for each slide. However, instead of asking the LLM to generate the content directly from the outline and the whole input document as the context, we ask it to associate each slide title to one or more sections of the document using the bird's-eye view of the document as generated in Section \ref{sec:bird's-eye}. This has the following advantages: (i) For each generated slide, we can attribute it to some specific sections (and subsections) of the document. This will make the content of the slide more reliable and make it easy for users to update it. (ii) Grounding a slide to some specific parts of the document reduces hallucinations \citep{yue2023automatic}. (iii) The flow of information in the presentation need not strictly follow the information flow in the given document. This non-linearity of the flow makes the generated presentation more similar to ones prepared by humans \citep{bartsch2003effectiveness}. (iv) We do not need to feed the entire document to the LLM, making it suitable for long documents. Appendix \ref{sec:appendixP_slides2sections} has the exact prompt.
%The exact prompt is shown in Appendix \ref{sec:appendixP_slides2sections}.

% \begin{boxA}
% Think step by step.
% \begin{enumerate}
%     \item You are given with the following titles:
%         <Outline>
%     \item And a list of keys:
%         <Headings present in the bird's-eye view>
%     \item Each key is associated with some text as presented below:
%         <Bird's-eye view>
%     \item The task is to find 1-2 significantly matched keys for each slide title
%     \item The matching should be done based on the similarity of the text associated with the keys with the given slide titles
% \end{enumerate}    
% \end{boxA}
% The output of the above call maps each slide title present into the outline to one or more sections of the document. 
Since the output of LLMs are probabilistic in nature and often verbose, we use edit distance \citep{navarro2001guided} to match each section title produced by the LLM with the ones present in the document. We select the section present in the document when the match is more than $90\%$. Using edit distance to select section titles from the document makes our system robust to any hallucination in the output produced by the LLM during this mapping. 

% Please note that this mapping between slide titles and the sections of the document can increase the reliability of the generated slides to the users. It attributes the source content and help them further edit the slides, which otherwise is a major problem in state-of-the-art LLMs due to their tendency to hallucinate \cite{yue2023automatic}. A sample output from this stage of our pipeline is shown below.

\subsection{Slide Text Generation}
% {\color{red}Sambaran}
Once we get the individual slide titles and the document sections (or subsections) associated to each slide, we target to generate the text content for each slide at a time. If there are multiple sections associated to a slide, we concatenate the content of those sections into a single one before feeding it to the LLM. However, generating the text independently for each slide may not ensure the natural flow of the presentation. Hence, to generate the content of the slide $L_k$, we feed the Slide Title $O_k$, concatenated text from the associated sections, along with the slide title and content of the previous slides $L_1,\cdots,L_{k-1}$, $\forall k = 2,\cdots,K$, to an LLM. The detailed prompt is mentioned in Appendix \ref{sec:appendixP_slideContent}.
% \begin{boxA}
% Think step by step.
% \begin{enumerate}
%     \item You have to generate the slide number {k}.
%     \item Previous slide headings and slide contents are given below as: $<L1,\cdots,L_{k-1}>$
%     \item The current slide heading and the source of the content is given at the end of this prompt.
%     \item The content of the slide is very relevant to the given slide title and brief
%     \item Ensure that this slide does not have any content repeated from the previous slides.
%     \item The flow of the overall presentation is nice.
%     \item Slide Title: $<O_k>$
%     \item Source of text: <concatenated text from the mapped sections>
% \end{enumerate}
% \end{boxA}
% 
The output of this stage generates a presentation with the slide titles and the corresponding text in the form of bullet points. We have ensured that the content of each slide is related to its title, maintains a good flow of information and concise in nature.

\section{Image Extraction}
Next, we aim to add images in the slides. We use a set of heuristics and a ranking algorithm based on the similarity of the text and images in a common subspace through a VLM.
The content extraction module outputs all the possible images present in a document which can even contain page boundary lines, small and repeated logo, large images with very bad aspect ratio to be shown in a slide, etc. Thus, we use simple rules to remove images with bad aspect ratio and repeated images from the set of candidate images to go into a slide.

Next, for each slide $L_k$, we use the output of Section \ref{sec:slide2sec} to get the sections $S_{ck}$ from the document that contributed to the slide.
% Then the set of all figures $F_c$ that are present in the same pages as $S_c$ are  taken. In case where the section information for a figure is not available, 
We use the positional information to consider only the figures $F_{ck}$ present within a  distance from the contributing sections in the document. After this, a suitability score $\alpha_{ck}$ of each figure $F_{ck}$ is computed as the cosine distance of the CLIP embedding \citep{radford2021learning} of $F_{ck}$ and the CLIP embedding of the text of slide $L_K$. Then the image with the highest $\alpha_{ck}$ is chosen for the slide $L_k$ subject to $\alpha_{ck} > \alpha_{min}$, where $\alpha_{min}$ is a threshold which we set as $80\%$.

\section{Experimental Evaluation}

\subsection{Experimental Setup and Baselines}
Our proposed approach DocPres does not need any training data since it is based on a combination of LLM and VLM (CLIP model). We choose GPT-3.5-turbo \citep{ouyang2022training} as the LLM, due to its known superior performance in many NLP tasks and its larger context length (which will be required by the baselines). 
%as the LLM for our approach and all the baselines as it has been applied on many NLP tasks such as summarization and question-answering and also has relatively longer contaext length which is a need for our baseline. 
We use the publicly available test split of SciDuet dataset \citep{sun2021d2s} which consists of 100 research papers from ICML and NeurIPS conferences as our input documents. 

We use the following four baselines:
(i) \textbf{D2S}: We use D2S \citep{sun2021d2s} as a semi-automatic baseline where the slide titles are taken from the ground truth slides from SciDuet dataset and the algorithm generated the content of the presentation.
(ii) \textbf{GPT-Flat}: Here, we feed the entire document to GPT-3.5-turbo and use a descriptive prompt to generate a presentation consisting of slide title and text in bullet points. 
(iii) \textbf{GPT-COT}: Instead of a descriptive prompt, we use chain-of-thought prompting in this baseline with GPT-3.5-turbo. 
(iv) \textbf{GPT-Cons}: We explicitly mention the maximum number of words in a bullet point and the number of bullet point in each slide with COT prompting. The detailed prompts are presented in Appendix \ref{sec:appendix_baselines}.

\subsection{Automated Evaluation Metrics}
There is no established evaluation framework exists for document to slides generation. We have carefully chosen three unsupervised metrics here:
(i) \textbf{Coverage}: This is to capture the semantic coverage of the content of the document in the presentation. We calculate coverage at paragraph-slide level and sentence-bullet level. To calculate coverage, we compute the average cosine similarity between a paragraph (or sentence) embeddings from the document and a slide (or bullet point) embedding from the generated presentation.
(ii) \textbf{Perplexity (PPL)}: It measures how likely the language model (GPT-2 here) is to generate the sequence. It is obtained using GPT-2, as discussed in \citet{liu-etal-2021-learning}. If GPT-2 assigns a high probability to the tokens present in a sentence, the perplexity will be lower, indicating a fluent and grammatically correct sentence.
(iii) \textbf{LLM-Eval}: Here, we use Mistral-7B-Instruct-v0.2 \citep{jiang2023mistral} to evaluate the overall presentation quality in terms of organization, effectiveness, clarity, coherence, and the ability to convey complex ideas to the audience. This metric is called LLM-Eval \cite{liu-etal-2023-g}. More details about the metrics can be found in the Appendix \ref{sec:appendix_metrics}. 
% We have not included any metric on images since all the GPT-based baselines are text only, and our CLIP based algorithm with filtering strategy is 

% \renewcommand{\arraystretch}{1.4}
% \setlength{\tabcolsep}{5pt}
\begin{table}[ht]
\centering
\resizebox{1\linewidth}{!}{
\begin{tabular}{c|cc|c|c}
    \toprule
      Method & \multicolumn{2}{c}{Coverage (\%) $\uparrow$} & PPL $\downarrow$ & LLM-Eval $\uparrow$ \\
      & Paragraph & Sentence & &\\
    \hline
    D2S & 38.48 $\pm$ 5.43 & 24.24 $\pm$ 3.38 & 77.38 $\pm$ 28.95 & 7.61 $\pm$ 1.05\\
    % \hline
    GPT-Flat & 33.41 $\pm$ 8.12 & 22.83 $\pm$ 4.03 & 133.51 $\pm$ 96.92 & 8.94 $\pm$ 0.36\\
    GPT-COT & 34.83 $\pm$ 6.06 & 23.38 $\pm$ 4.07 & 104.14 $\pm$ 53.70 & \textbf{8.98 $\pm$ 0.26} \\
    GPT-Cons & 34.59 $\pm$ 7.63	& 23.31 $\pm$ 4.17 & 121.37 $\pm$ 112.16 & 8.90 $\pm$ 0.33\\
    \hline
    \textbf{DocPres} & \textbf{39.13 $\pm$ 5.68} & \textbf{24.73 $\pm$ 3.48} & \textbf{58.01 $\pm$ 20.44} & 8.95 $\pm$ 0.32\\
    \bottomrule
\end{tabular}
}
\caption{Results with different automated metrics}
\label{tab:automated_results}
\end{table}

\subsection{Results and Analysis}
Table \ref{tab:automated_results} compares the performance of DocPres with the baselines. Please note that D2S has some advantage on SciDuet dataset since it was specifically trained on the same dataset where all other algorithms including DocPres are LLM-based.
% Interestingly, DocPres is able to achieve better results on all the metrics, except LLM-Eval for which all the LLM-based approaches perform very close to each other.
Interestingly, DocPres performs the best among the baselines for Coverage and PPL, where the margin is significant compared to other LLM based approaches.
For LLM-Eval, all the LLM-based approaches perform very close to each other.
% {\color{red}Even in LLM-Eval, DocPres has a very high score, higher than all baselines but one.} 
This specifically supports our hypothesis that dividing a complex task into smaller sub-tasks and providing limited context for each sub-task helps to improve the overall performance of an LLM compared to solving the task directly with a very long context.

\subsection{Human Evaluation} \label{sec:human_evaluation}
We have also conducted a small scale human survey to understand the quality of the generated presentation to human experts. First, we selected five research papers from ACL workshops which are relatively easy to follow. We hired~\footnote{\url{https://www.upwork.com/}} two professional reviewers who have reasonable understanding of NLP and have good presentation generation skill. Based on our discussion with subject matter experts, we decided the following criteria to evaluate the quality of a generated presentation from a given document: (1) Readability: \textit{How good is the language and readability?}, (2) Consistency: \textit{Is a slide title consistent with the slide content?}, (3) Coverage: \textit{Does the presentation cover all important parts of the document?}, (4) Diversity: \textit{Is the content of the presentation non-repetitive enough?}, (5) Flow: \textit{How is the flow of information in the presentation?} and (6) Usability: \textit{Is the generated presentation good enough for an initial draft?}.
The evaluators are instructed to score a generated presentation against each of these metrics in a scale of 1 (lowest in quality) to 5 (best in quality)~\footnote{We could not use D2S in this study since we were not able to run its publicly available code on any other dataset except SciDuet.}.

Human evaluation results in Table \ref{tab:human_rsults_research} shows that the slides generated by DocPres are consistently rated high by human experts with a good margin compared to the baselines. Interestingly, the scores of all direct GPT-based baselines are very close to each other showing that different prompting techniques could not generate visible difference in the generated presentation. The reviewer appreciated the output of DocPres from different perspectives such as \textit{"The language and grammar are all fine"}, \textit{"The main text and the slide title are closely related"}, \textit{"The flow is good"}, etc. However, there were a few concerns such as DocPres
\textit{"Covers a lot of content from the PDF but does not deep dive"} and \textit{"The deck keeps on repeating the benefits of text mining"}. 
We also asked reviewers to comment on the images extracted by DocPres. Reviewers consistently appreciated the precision of the selected images (because of our filtering strategy), however complained about the missing images. This is because research papers have many non-natural images which CLIP based algorithm fails to understand.
Overall, the reviewers agree that compared to the baselines, the generated presentations from DocPres can serve well as an initial draft.

\begin{table}[ht]
\centering
\resizebox{1.1\linewidth}{!}{
\begin{tabular}{c|cccccc}
    \toprule
      Method & Readability & Consistency & Coverage & Diversity & Flow & Usability \\
    \midrule
    GPT-Flat & 2.30 $\pm$ 1.16 & 2.20 $\pm$ 1.13 & 1.30 $\pm$ 0.48 & 2.80 $\pm$ 1.54 & 1.70 $\pm$ 0.67 & 1.20 $\pm$ 0.63 \\
    GPT-COT & 2.30 $\pm$ 1.16 & 2.40 $\pm$ 1.35 & 1.50 $\pm$ 0.85 & 2.80 $\pm$ 1.54 & 1.70 $\pm$ 0.67 & 1.20 $\pm$ 0.63 \\
    GPT-Cons & 2.30 $\pm$ 1.16 & 2.00 $\pm$ 1.05 & 1.10 $\pm$ 0.31 & 2.80 $\pm$ 1.54 & 1.70 $\pm$ 0.67 & 1.20 $\pm$ 0.63 \\
    \midrule
    \textbf{DocPres} & \textbf{3.90 $\pm$ 0.73}  & \textbf{3.80 $\pm$ 1.39} & \textbf{2.70 $\pm$ 1.16} & \textbf{2.90 $\pm$ 1.44} & \textbf{2.70 $\pm$ 0.82} & \textbf{3.20 $\pm$ 1.22} \\
    \bottomrule
\end{tabular}
}
\caption{Results of human evaluation}
\label{tab:human_rsults_research}
\end{table}

\section{Discussions and Conclusion}
This work presented a novel multi-staged framework for generating presentations from documents. By breaking down the task into five sub-tasks, our approach achieved significant improvements compared to baselines and single-shot prompting to LLMs. Comprehensive evaluations, both automatic and human, confirmed the superiority of our multi-stage approach. The presentations from our approach demonstrated better coverage, readability, consistency, diversity, flow, and overall usability. The success of our multi-stage approach highlights the benefits of decomposing complex tasks into smaller and well-defined subtasks for LLMs. 

\section{Limitations}
Following are some of the limitations of our current work: \\
1. Our image selection approach suffers the limitations of CLIP. While clip is trained on datasets dominated by images that convey naturally occurring items like photographs and cartoons, documents contain a lot of images that are not naturally occurring items like illustrations, flow charts and graphs. 
\\
2.While we are yet to analyse the computational cost of our methodology, we believe there is opportunity to reduce cost as there is a lot of LLM usage in our methodology\\
3. Our method currently works by converting a single document into a presentation. While this may be the case in many academic presentations where a research paper needs to be converted into a presentation, often times information needs to come from multiple documents and made into a slide in other types of presentations. We are currently not handling such a scenario.

\bibliography{custom}

\clearpage

\appendix

\section*{Appendix}

\section{Prompt to Generate an Outline}
\label{sec:appendix_outline}
Table \ref{tab:prompt_outline} shows the prompt that we used for generating the outline of the presentation.
\renewcommand{\arraystretch}{1.2}
\setlength{\tabcolsep}{4pt}
\begin{table}
\centering \scriptsize
\begin{tabular}{|l|}
\hline
\makecell[l]{From the following text which contains a set of headings and some \\content within each heading:\\\\TEXT\\\\Extract the most important headings present in it.\\Reduce the length of each heading to five words if they are lengthy.
}\\
\hline
\end{tabular}
\caption{Prompt to generate an outline.}
\label{tab:prompt_outline}
\end{table}

\section{Prompt to Map Slides to Sections}
\label{sec:appendixP_slides2sections}
Table \ref{tab:map_slide_2_sections} shows the prompt that we used for mapping slides to sections.

\renewcommand{\arraystretch}{1.2}
\setlength{\tabcolsep}{4pt}
\begin{table}
\centering \scriptsize
\begin{tabular}{|l|}
\hline
\makecell[l]{
Think step by step\\\\You are given with the following title:\\ \{outline_headings\}\\\\and a list of keys:\\\{document_heading_from_bird_eye_view\}\\\\Each key is associated with some text as presented in the dictionary format \\below:\\\{bird_eye_view\}\\\\The task is to find 1-2 significantly matched keys. The matching should be \\done based on the similarity of the text associated with the keys with the \\given heading.\\Matching keys are: <semicolon separated list if more than a single key>
}\\
\hline
\end{tabular}
\caption{Prompt to map slides to section.}
\label{tab:map_slide_2_sections}
\end{table}

% \begin{boxA}
% Think step by step.
% \begin{enumerate}
%     \item You are given with the following titles:
%         <Outline>
%     \item And a list of keys:
%         <Headings present in the bird's-eye view>
%     \item Each key is associated with some text as presented below:
%         <Bird's-eye view>
%     \item The task is to find 1-2 significantly matched keys for each slide title
%     \item The matching should be done based on the similarity of the text associated with the keys with the given slide titles
% \end{enumerate}    
% \end{boxA}
% The output of the above call maps each slide title present into the outline to one or more sections of the document.

% A sample output from this stage is shown below.
% \begin{boxA}
%     <Slide 1> $\xleftrightarrow{}$ <Doc Section 1> <Doc Section 5>\\
%     <Slide 2> $\xleftrightarrow{}$ <Doc Section 3> <Doc Section 5>\\
%     $\cdots$
% \end{boxA}

\section{Prompt to Generate Slide Content}
\label{sec:appendixP_slideContent}

Table \ref{tab:generate_slide} shows the prompt that we used for generating the slide content.

\renewcommand{\arraystretch}{1.2}
\setlength{\tabcolsep}{4pt}
\begin{table}
\centering \scriptsize
\begin{tabular}{|l|}
\hline
\makecell[l]{
You are a presentation generator from a source of text. You have to generate\\the slide number \{slide_index\}.\\Previous slide headings and slide contents are given below in the format of a\\list of dictionaries.\\\{previous_slide\}\\Given the following slide heading and the source of text respectively, create\\the content of the slide number \{slide_index\} such that:\\1. The slide should have maximum {max_bullet} bullet points.\\2. Ensure that the content of the bullet points are coming strictly from the\\given source of text only.\\3. The content of the slide is very relevant to the given slide heading\\4. Each bullet point should have a maximum of 10 words\\5. Ensure that this slide does not have any content repeated from the\\previous slides.\\6. The flow of the overall presentation is nice.\\7. Do not prefix the slide title before the bullet points in the output\\\\Slide Title: HEADING\\\\Source of text: TEXT
}\\
\hline
\end{tabular}
\caption{Prompt to generate slide.}
\label{tab:generate_slide}
\end{table}

% \begin{boxA}
% Think step by step.
% \begin{enumerate}
%     \item You have to generate the slide number {k}.
%     \item Previous slide headings and slide contents are given below as: $<L1,\cdots,L_{k-1}>$
%     \item The current slide heading and the source of the content is given at the end of this prompt.
%     \item The content of the slide is very relevant to the given slide title and brief
%     \item Ensure that this slide does not have any content repeated from the previous slides.
%     \item The flow of the overall presentation is nice.
%     \item Slide Title: $<O_k>$
%     \item Source of text: <concatenated text from the mapped sections>
% \end{enumerate}
% \end{boxA}

\section{Prompt for the Baselines}
\label{sec:appendix_baselines}
\subsection{Prompt for GPT-Flat}
\label{sec:appendix_gpt_flat}
Table \ref{tab:gpt_flat_prompt} shows the prompt for GPT-Flat baseline.

\renewcommand{\arraystretch}{1.2}
\setlength{\tabcolsep}{4pt}
\begin{table}
\centering \scriptsize
\begin{tabular}{|l|}
\hline
\makecell[l]{You're an AI assistant that will help create a presentation from a document. \\You will be given section heading and paragraphs in that section. Your task \\is to create a presentation with ONLY \#\#number_of_slides\#\# slides from \\the document. For every slide, output the slide title and bullet points in the \\slides. 
Please follow the following structure in the output. Do not \\output slide number. \\
Slide Title: The slide title  \\
Bullet Points:  \\
New line separated bullet points  \\ \\
Following is the document, which contains section heading and paragraphs \\under that heading. \\
----------Document Started---------- \\
\#\#document\#\# \\
----------Document Ended---------- \\ \\
Presentation (only \#\#number_of_slides\#\# slides): }\\
\hline
\end{tabular}
\caption{Prompt for GPT-Flat}
\label{tab:gpt_flat_prompt}
\end{table}

\subsection{Prompt for GPT-COT}
\label{sec:appendix_gpt_cot}
Table \ref{tab:gpt_cot_prompt} shows the prompt for GPT-COT baseline.

\renewcommand{\arraystretch}{1.2}
\setlength{\tabcolsep}{4pt}
\begin{table}
\centering \scriptsize
\begin{tabular}{|l|}
\hline
\makecell[l]{You're an AI assistant that will help create a presentation from a document. You \\will be given section heading and paragraphs in that section. Your task is to create \\a presentation with ONLY \#\#number_of_slides\#\# slides from the document. For \\every slide, output the slide title and bullet points in the slides. Please follow the\\ steps provided below. \\
1. Begin by thoroughly reading and understanding the document. Identify the \\main points, key messages, and supporting details. \\
2. Find relations between different paragraphs that could be presented in the \\same slide. \\
3. Create a high-level outline for your presentation. Identify the main sections or \\topics that you'll cover. This will serve as the skeleton for your slides. \\
4. Choose the most important information from the document to include in your \\presentation. Focus on key messages and supporting details that align with your \\presentation objectives. \\
5. Organize the selected content into slides, maintaining a logical flow. Each \\slide should represent a clear point or topic, and the overall structure should make \\sense to your audience. \\
6. Make sure slides are descriptive. \\
7. Presentation should have only \#\#number_of_slides\#\# slides. \\
8. Please follow the following structure. Do not output slide number.\\  
Slide Title: The slide title \\
Bullet Points: \\
New line separated bullet points \\ \\
Following is the document, which contains section heading and paragraphs under \\that heading.\\ 
----------Document Started---------- \\
\#\#document\#\# \\
----------Document Ended----------\\ \\ 
Presentation:  }\\
\hline
\end{tabular}
\caption{Prompt for GPT-COT.}
\label{tab:gpt_cot_prompt}
\end{table}

\subsection{Prompt for GPT-Cons}
\label{sec:appendix_gpt_cons}
Table \ref{tab:gpt_cons_prompt} shows the prompt for GPT-Cons baseline.

\renewcommand{\arraystretch}{1.2}
\setlength{\tabcolsep}{4pt}
\begin{table}
\centering \scriptsize
\begin{tabular}{|l|}
\hline
\makecell[l]{You're an AI assistant that will help create a presentation from a document.\\You will be given section heading and paragraphs in that section. Your task\\is to create a presentation with ONLY \#\#number_of_slides\#\# slides from the\\document. For every slide, output the slide title and bullet points in the slides. \\Please follow the steps provided below. \\
1. Begin by thoroughly reading and understanding the document. Identify the \\main points, key messages, and supporting details. \\
2. Find relations between different paragraphs that could be presented in the \\same slide. \\
3. Create a high-level outline for your presentation. Identify the main sections\\or topics that you'll cover. This will serve as the skeleton for your slides. \\
4. Choose the most important information from the document to include in \\your presentation. Focus on key messages and supporting details that align\\with your presentation objectives. \\
5. Organize the selected content into slides, maintaining a logical flow. Each \\slide should represent a clear point or topic, and the overall structure should\\make sense to your audience. \\
6. Make sure slides are descriptive. \\
7. Presentation should have only \#\#number_of_slides\#\# slides. \\
8. Each slide should have around 7 bullet points. Each bullet point should\\have around 15 words.\\
9. Please follow the following structure. Do not output slide number.\\  
Slide Title: The slide title \\
Bullet Points: \\
New line separated bullet points \\ \\
Following is the document, which contains section heading and paragraphs\\under that heading.\\ 
----------Document Started---------- \\
\#\#document\#\# \\
----------Document Ended----------\\ \\ 
Presentation:  }\\
\hline
\end{tabular}
\caption{Prompt for GPT-Cons}
\label{tab:gpt_cons_prompt}
\end{table}

\section{Details on Automated Evaluation Metrics}\label{sec:appendix_metrics}
1. \textbf{Coverage}: It is an unsupervised metric which intuitively capture how much does a subset ``cover'' the content of the super set. In literature, it has been used for extractive summarization \citep{kothawade2020deep}. We use the following definition of Coverage (at \textbf{paragraph} to slide level) in this work:
\[ Coverage = \frac{\sum_{\mathbf{e}_p \in D}\sum_{\mathbf{e}_s \in P}cosine\bigl(\mathbf{e}_p, \mathbf{e}_s \bigr)}{|D||P|} \times 100 \%\]
Here, $\mathbf{e}_p$ is a paragraph embedding from the given document and $\mathbf{e}_s$ is a slide embedding from the generated presentation as obtained by a sentence transformer model \cite{reimers-2019-sentence-bert}.
%If some paragraph is similar to multiple slides, it is covered well in the overall presentation. 
Similarly, coverage can also be computed ta \textbf{sentence} level by replacing a paragraph with a sentence from the  document and a slide with a bullet point (or sentence) from the presentation in the equation above. Sentence level coverage offers a finer granularity than paragraph-level coverage. More is the Coverage, better is the presentation.\\
% 3. \textbf{Perplexity}: Perplexity is a key metric to indicate the fluency and grammatical correctness of the generated text. We use GPT-2 to get the perplexity of the sentence, as discussed in \citet{liu-etal-2021-learning}.\\
2. \textbf{Perplexity (PPL)}: Perplexity is a metric to indicate the fluency of the generated text. It is obtained using GPT-2, as discussed in \citet{liu-etal-2021-learning}. Perplexity measures how likely the language model (GPT-2 here) is to generate the sequence. If GPT-2 assigns a high probability to the token present in the sentence, the perplexity will be lower, indicating a fluent and grammatically correct sentence.\\
3. \textbf{LLM-Eval for presentation quality}: G-Eval \cite{liu-etal-2023-g} is a well-established metric that uses GPT to evaluate various NLP tasks. It has a very high correlation with humans. We believe that G-Eval might be biased to GPT output, so instead of GPT, we use open-source LLMs (Mistral-7B-Instruct-v0.2). We call this metric LLM-Eval. We use LLM-Eval to measure the overall presentation quality in terms of organization, effectiveness, clarity, coherence, and the ability to convey complex ideas to the audience. Table \ref{tab:g_eval_prompt} shows the prompt we used for LLM-Eval to evaluate the presentation quality.

\renewcommand{\arraystretch}{1.2}
\setlength{\tabcolsep}{4pt}
\begin{table}
\centering \scriptsize
\begin{tabular}{|l|}
\hline
\makecell[l]{On a scale of 0-10, rate the effectiveness, clarity, and overall quality of \\the following text presentation, considering factors such as organization,\\ coherence, and the ability to convey complex ideas to the audience. \\0 is the lowest score, whereas 10 is the highest score.\\ \\ Presentation:\\\#\#presentation\#\#\\ \\Score (an integer between 0 and 10):}\\
\hline
\end{tabular}
\caption{Prompt for LLM-Eval to evaluate the final presentation quality.}
\label{tab:g_eval_prompt}
\end{table}

% \section{Generated Presentations for Human Evaluation}
% In the following anonymous link, one can get the documents and the corresponding generated presentations by DocPres and the baseline algorithms used in the human evaluation as discussed in Section \ref{sec:human_evaluation}: \url{https://www.dropbox.com/scl/fo/egwq88zcmofm5rkwpuge6/h?rlkey=o9o3fyu0mbbwpfrehdm1qxz35&dl=0}.

\end{document}